\documentclass[]{bytedance_seed}

\usepackage[toc,page,header]{appendix}
\usepackage[utf8]{inputenc} 
\usepackage[T1]{fontenc}    
\usepackage{hyperref}       
\usepackage{url}            
\usepackage{booktabs}       
\usepackage{amsfonts}       
\usepackage{nicefrac}       
\usepackage{microtype}      
\usepackage{xcolor}         
\usepackage{amsmath} 

\usepackage{minitoc}

\title{A Unified Pairwise Framework for RLHF: Bridging Generative Reward Modeling and  Policy Optimization}

\author[*]{Wenyuan Xu}
\author[*]{Xiaochen Zuo}
\author[]{Chao Xin}
\author[]{Yu Yue}
\author[\dagger]{Lin Yan}
\author[]{Yonghui Wu}

\affiliation[]{ByteDance Seed}

\contribution[*]{Co-first authors}
\contribution[\dagger]{Corresponding authors}

\abstract{
Reinforcement Learning from Human Feedback (RLHF) has emerged as a important paradigm for aligning large language models (LLMs) with human preferences during post-training. This framework typically involves two stages: first, training a reward model on human preference data, followed by optimizing the language model using reinforcement learning algorithms. However, current RLHF approaches may constrained by two limitations. First, existing RLHF frameworks often rely on Bradley-Terry models to assign scalar rewards based on pairwise comparisons of individual responses. However, this approach imposes significant challenges on reward model (RM), as the inherent variability in prompt-response pairs across different contexts demands robust calibration capabilities from the RM. Second, reward models are typically initialized from generative foundation models—such as pre-trained or supervised fine-tuned models, despite the fact that reward models perform discriminative tasks, creating a mismatch. This paper introduces Pairwise-RL, a RLHF framework that addresses these challenges through a combination of generative reward modeling and a pairwise proximal policy optimization (PPO) algorithm. Pairwise-RL unifies reward model training and its application during reinforcement learning within a consistent pairwise paradigm, leveraging generative modeling techniques to enhance reward model performance and score calibration. Experimental evaluations demonstrate that Pairwise-RL outperforms traditional RLHF frameworks across both internal evaluation datasets and standard public benchmarks, underscoring its effectiveness in improving alignment and model behavior.
}

\date{\today}
\correspondence{Lin Yan at \email{neil@bytedance.com}}

\begin{document}
\maketitle


\section{Introduction}

Large language models have demonstrated unprecedented capabilities in generating human-like text. However, aligning these models with human values and preferences—ensuring they produce safe, helpful, and contextually appropriate outputs—remains a critical challenge. RLHF has emerged as the dominant framework for post-training alignment, enabling models to learn from human preferences rather than predefined rules. RLHF typically operates in two stages: first, training a reward model to predict human preferences from pairwise comparisons of model outputs, and second, optimizing the language model using reinforcement learning algorithms that leverage these rewards. Despite its success, current RLHF approaches face some limitations. 

The first challenge lies in the mismatch between the pairwise nature of human preference data and the scalar reward signals required for reinforcement learning. Traditional RLHF relies on Bradley-Terry models to convert pairwise comparisons into scalar rewards, but these rewards often lack calibration across diverse prompts and responses. This calibration issue can lead to unstable training dynamics and suboptimal alignment, as the RL algorithm may misinterpret the magnitude of rewards in different contexts. The second limitation stems from the initialization of reward models. Most RLHF pipelines initialize reward models from generative foundation models, such as pre-trained or supervised fine-tuned LLMs. However, reward models perform a discriminative task—ranking outputs based on human preferences—while generative models are optimized for sequence generation. This architectural and objective mismatch can hinder the reward model’s ability to accurately capture human preferences, propagating errors into the subsequent RL stage. 

To address these challenges, we introduce Pairwise-RL, a RLHF framework that unifies reward model training and reinforcement learning within a consistent pairwise paradigm. Pairwise-RL tackles the calibration problem by leveraging generative modeling techniques to enhance reward model performance and ensure consistent scoring across diverse scenarios. In this paper, we make the following contributions: 
\begin{itemize}
\item We introduce a pairwise PPO algorithm that operates directly on pairwise comparisons, bypassing the limitations of scalar reward approximations. 
\item We leverage generative modeling techniques to enhance reward model calibration and performance, and apply it to RLHF training.
\item Experimental evaluations on internal datasets and public benchmarks demonstrate that Pairwise-RL outperforms traditional RLHF methods, achieving better alignment with human preferences and improving model behavior across a range of tasks. 
\end{itemize}

\section{Preliminary}
The reward model plays a pivotal role in translating human preferences into actionable training signals for large language models. Traditional RLHF pipelines typically train a RM to assign scalar rewards to model outputs by fitting a Bradley-Terry model on pairwise human preference comparisons. The probability that one response is preferred by another is often given by:
\begin{equation}
P(y_1 \succ y_2|q)=\sigma(r(q,y_1)-r(q,y_2))
\label{eq:rm_score}
\end{equation}
where $q$ represents a prompt sampled from the dataset $D$, $y_1,y_2$ stands for corresponding responses.

The RLHF task which aiming to maximize rewards given by the reward model, usually under a KL-constrain, is defined as:
\begin{align}\label{eq:objective}
   \pi^* =  \arg\max_\pi \mathbb{E}_{\pi, q \sim D} \left[ \sum_{t = 0}^T  \left(r(s_t, a_t)-\beta \text{KL} \big( \pi(\cdot | s_t) \| \pi_{\text{ref}}(\cdot | s_t) \big)\right) \right]
\end{align}
Here, $T$ denotes the total number of decision steps; $r(s_t, a_t)$ is the token-level reward provided by the reward function; $\beta$ serves as a coefficient controlling the strength of the KL-regularization; and $\pi_{\text{ref}}$ denotes the initialization policy.

One commonly used algorithm to optimize this objective is PPO\cite{ppo}, which aligns well with the constrained optimization framework inherent in the RLHF setup by ensuring stable policy updates through its clipped surrogate objective. PPO employs a clipped surrogate objective to update the policy, with the core principle of constraining policy changes during each update to avoid destabilizing large modifications. Let $\pi_{\theta}(a|s)$ denote the policy parameterized by $\theta$, and $\pi_{\theta_{\text{old}}}(a|s)$ represent the policy from the previous iteration. The surrogate objective function for PPO is defined as:
\begin{equation}
\mathcal{L}^{CLIP}(\theta)=\hat{\mathbb{E}}_t\left[\min\left(r_t(\theta)\hat{A}_t,\text{clip}(r_t(\theta), 1-\epsilon, 1+\epsilon)\hat{A}_t\right)\right]
\end{equation}
where $r_t(\theta)=\frac{\pi_{\theta}(a_t|s_t)}{\pi_{\theta_{\text{old}}}(a_t|s_t)}$ is the probability ratio, $\hat{A}_t$ is the estimated advantage at time step $t$, and $\epsilon$ is a hyperparameter controlling the clipping range.

Generalized Advantage Estimation (GAE) is utilized in PPO to compute more accurate advantage estimates by integrating multi-step bootstrapping, thereby reducing variance\cite{gae}. For a trajectory of length $T$, the advantage estimate $\hat{A}_t$ at time step $t$ is calculated as:
\begin{equation}
\hat{A}_t=\sum_{l = 0}^{T-t-1}(\gamma\lambda)^l\delta_{t + l},
\label{eq:gae_definition}
\end{equation}
where $\gamma$ is the discount factor, $\lambda\in[0, 1]$ is the GAE parameter, and $\delta_t=r_t+\gamma V(s_{t + 1})-V(s_t)$ is the temporal-difference (TD) error. Here, $r_t$ is the reward at time step $t$, and $V(s)$ is the value function. Notably, since a discount factor of $\gamma = 1.0$ is conventionally adopted in RLHF, we omit $\gamma$ from subsequent notation for simplicity.

\section{Related Work}

\subsection{Pairwise reward models}
The proposed framework introduces a unified pairwise RLHF training approach, witch include pairwise reward model and pairwise PPO. Several studies have investigated alternatives to traditional Bradley-Terry (BT) models, such as the pairwise generative reward model\citep{grm}. Generative reward models often process pairs of responses within a shared context, enabling the model to directly evaluate their relative quality. A key distinction among these approaches lies in whether the model generates a binary preference decision directly or incorporates intermediate reasoning steps, such as Chain of Thought (CoT).

Traditional generative reward models may produce a single token (e.g., "better" or "worse") to indicate preference, relying on the model’s implicit understanding of the response pair. However, recent advancements introduce CoT-based methods, where the model first generates a rationale for its preference before rendering a final judgment. This approach leverages the model’s reasoning capacity to align decisions with human-like logic, but it also introduces variability due to the stochastic nature of CoT generation. To mitigate this, some frameworks employ voting strategies, aggregating multiple CoT-driven judgments to estimate the model’s average preference. These strategies aim to stabilize the reward signal by accounting for the uncertainty in intermediate reasoning steps, thereby improving the consistency of pairwise comparisons.

\subsection{Pairwise reinforcement learning methods}
Some recent related works have also attempted to use the pairwise approach to deal with RLHF. Direct Preference Optimization (DPO) directly optimize the policy to maximize the likelihood of preferred responses over non-preferred ones \cite{dpo}, eliminating the need for an explicit reward model or reinforcement learning by leveraging a mathematical equivalence between the reward and policy optimization objectives. A novel trajectory-wise policy gradient algorithm P3O is proposed \cite{p3o}, which directly operates on comparative rewards to align large language models with human preferences. Both of these two methods approach language generation as Contextual Bandit and view the whole trajectory as a single actor. Compared other approaches that treats each individual token as an actor (E.g. \cite{ppo} \cite{a3c} \cite{sac}), the main drawbacks of this method are that the rewards are overly sparse, making it difficult to determine the contributions of important tokens for credit assignment, and the sample efficiency is relatively low. In addition, a reward shaping technique Preference As Reward (PAR) is proposed to obtain the centered reward by sigmoid function \cite{par}. And robust RLHF designs a contrastive reward using offline sampled baseline responses to improve model robustness to noise in the reward function \cite{robustRLHF}. These two works consider introducing a baseline when using the reward model to mitigate the reward hacking problem, demonstrating benefits of using pairwise rewards in the RL process.

\section{Approach}

\subsection{Pair-wise Reward Model}
In the proposed framework, the reward model is designed to operate under a pairwise comparison paradigm, which directly aligns with the RL objective of maximizing the probability that a generated response outperforms its ground-truth anchor. Unlike traditional approaches where the RM assigns absolute scores to individual responses, we adopt pairwise RM that evaluates two responses (the generated output and its anchor) jointly. This design simplifies the RM’s task by reducing it to a relative judgment rather than an absolute scoring problem. By avoiding the need to calibrate absolute reward magnitudes, the pairwise RM mitigates the generalization challenges under distributional shifts, as its relative judgment framework reduces sensitivity to absolute score calibrations that may fail when data distributions change. 

To fully leverage the capabilities of pretrained language models, we structure the pairwise comparison task as a natural language understanding problem. Specifically, we format each comparison instance as a textual query that explicitly asks whether the generated response $y$ is preferable to its ground-truth anchor $y*$ under rule $R$ and question $q$. For example, the input to the reward model is constructed as "Given the question $q$ and the rule $R$, is the response $y*$ better than $y$? Answer yes or no. [MASK]" where the [MASK] token is replaced by "yes" or "no" during training. Under normal circumstances, $R$ will be left empty, unless the judgment of this prompt requires special rules.

By framing the task in this manner, the reward model inherits the contextual understanding and semantic generalization abilities of the pretrained model, enabling it to perform discriminative judgments using the same architectural components that power its generative capabilities. The reward model is trained using a softmax cross-entropy loss over the vocabulary. Formally, for a batch of comparison instances \(\{y_i, y_i^*\}\), the cross-entropy loss is: 
\begin{equation}
\mathcal{L}_{\text{ce}} = -\frac{1}{N} \sum_{i=1}^N \log p_{\theta}(\text{yes/no} \mid q_i,R_i,y_i \succ y_i^*),
\label{eq:rm_cross_entr}
\end{equation}
where \(p_{\theta}\) denotes the probability distribution induced by the reward model's parameters \(\theta\). 

However, initializing the reward model with pretrained weights often introduces significant position bias, where the model may favor responses based on their positional order (e.g., preferring the first response in a pair) rather than their intrinsic quality. To address this, we employ a data augmentation strategy: for each training instance \((q, y, y^*, R)\), we generate two augmented samples: one where the generated response \(y\) precedes the anchor \(y^*\) in the input prompt, and another where their positions are swapped. By ensuring both variants are included in the same training batch, the model is compelled to focus on semantic content rather than positional heuristics. To further mitigate position bias, we introduce a mean squared error (MSE) constraint that penalizes inconsistencies in relative judgments. This constraint, formulated as: 
\begin{equation}
\mathcal{L}_{\text{pos}} = \frac{1}{N} \sum_{i=1}^N \left( p_{\theta}(\text{yes} \mid q_i,R_i,y_i \succ y_i^*) - p_{\theta}(\text{no} \mid q_i,R_i,y_i^* \succ y_i) \right)^2,
\label{eq:position_loss}
\end{equation}
encourages the reward model to produce symmetric judgments when the comparison order is reversed. The total training objective is a weighted combination of these two losses: 
\begin{equation}
\mathcal{L}_{rm} = \mathcal{L}_{\text{ce}} + \zeta \mathcal{L}_{\text{pos}},
\label{eq:rm_final_loss}
\end{equation}
where \(\zeta\) balances the trade-off between classification accuracy and positional consistency. $\zeta$ is relatively small to preserve some difference between different response positions.

\subsection{Pairwise Proximal Policy Optimization}
\label{qrl}
In this subsection, we will provide a detailed explanation of how to train the policy model using pairwise PPO based on a pairwise generative reward model. For a given prompt $q$, the ground truth answer $y^*$, and the online sampled answer $y\sim\pi_{\phi}$, we obtain the reward from the generative reward models as follows:
\begin{equation}
\label{rep}
\begin{aligned}
    r(y\mid y*,q) = p(y \succ y^* \mid q) =\frac{1}{2} \left(p(\text{yes} \mid q, y \succ y^*) + p(\text{no} \mid q, y^* \succ y) \right).
\end{aligned}
\end{equation}
Where $p(\text{yes} \mid q, y \succ y^*)$ represents the probability that $y$ is better than $y^*$, and $p(\text{no} \mid q, y^* \succ y)$ represents the probability that $y^*$ is worse than $y$, the average of these two probabilities is used as the final reward for $y$ ($R$ is not taken into account for the sake of conciseness). This approach offers two advantages: 1) It mitigates the bias of language models in predicting the words "yes" and "no". This approach guarantees that when \( y \) is identical to \( y^* \), the win rate is exactly 0.5, as the symmetric evaluations cancel out any positional artifacts; 2) It works like model ensemble to reduce the variance of individual estimations.

 In traditional RLHF, the optimization objective is to maximize the expected scalar reward \( r \) assigned by the reward model. The scalar reward \( r \) is often derived from absolute judgments, which can lead to unstable dynamics due to poor calibration across diverse prompts. However, in the Pairwise-RL framework, this objective is redefined to maximize the win probability of the generated response \( y \) over the ground-truth anchor \( y^* \) for a given prompt \( q \). The optimization objective in Pairwise-RL is formulated as: 
 \begin{align}\label{eq:pairwise_objective} \pi^* = \arg\max_\pi \mathbb{E}_{\pi, q \sim D}   \left( p(y \succ y^* \mid q) - \beta \cdot \text{KL} \big( \pi(\cdot \mid s_t) \| \pi_{\text{ref}}(\cdot \mid s_t) \big) \right) , \end{align} 
 this win probability is modeled as:
\begin{equation}p(y \succ y^* \mid q) =\begin{cases}
\text{softmax}(\text{logit}_{\text{yes}}) \approx \sigma(\text{logit}_\text{yes}-\text{logit}_\text{no}), & \text{pairwise RM} \\
\sigma(r_y-r_{y*}), & \text{pointwise RM}
\end{cases}
\label{eq:pairwise_p}
\end{equation}
 where \( \sigma(\cdot) \) denotes the sigmoid function, this transformation introduces a critical property: when the reward difference becomes large in magnitude, the sigmoid function saturates, compressing the win probability toward 0 or 1. Consequently, the advantage estimate \(\hat{A}_t\) in the PPO objective, which scales with the reward signal, diminishes in these saturated regions. This saturation effect acts as an implicit weighting mechanism across prompts. For prompts where the generated response \( y \) already significantly outperforms the anchor \( y^* \), the saturated win probability reduces the effective learning signal, preventing the policy from over-optimizing on these instances. Conversely, for prompts where \( y \) and \( y^* \) are closely matched, the win probability remains sensitive to small reward differences, amplifying the learning signal. This dynamic weighting mitigates the risk of "reward hacking", a phenomenon where models exploit flawed reward functions by disproportionately focusing on high-reward but contextually irrelevant patterns. Furthermore, clipping excessively large advantages serves to reduce the upper bound of the KL divergence between the policy before and after the update on these samples, thereby controlling the step size for this portion of the data, we will show this in the appendix.

On the other hand, the reward model compare the current response to the ground truth $y^*$ when estimating the reward, but the policy model can not access to $y^*$ during the sampling process, the learning of the value model will inevitably be misaligned with one of them. If the value model does not refer to $y^*$, it is difficult to obtain an accurately estimated value. If the value model also refers to $y^*$, it will be inconsistent with the policy model. In such a scenario, even if an accurate value signal is provided, there will still be a deviation between the update direction of the policy model and the true optimal direction. In fact, we have made attempts on both of these two learning methods of the value model.

\textbf{Value estimation with the ground truth (Value w/ GT)}

The first attempt is to let the value model also refers to the ground truth during the value estimation process. In implementation, the input of the value model changes from $q + y$ to $q + y^* + y$, which means the content of ground truth is visible during value estimation. The advantage of this approach is that the information input to the value model is completely aligned with that of the reward model, making it easier to learn accurate estimated values. However, the main problem with this approach is that, at the same timestamp of the same trajectory, the state evaluated by the value model is inconsistent with the state of the policy model when it executes the next action. Denote the state of the policy model at timestamp $t$ as $s_t$, and the state actually evaluated by the value model as $s'_t$. According to the Bellman Equation $v_\pi\left(s_t\right)=\sum_a \pi\left(a \mid s_t\right)\sum_{s_{t+1}, r}p\left(s_{t+1}, r|s_t, a\right)\left[r+\gamma v_\pi\left(s_{t+1}\right)\right]$, on the premise that the transition probability $p\left(s_{t+1}, r|s_t, a\right)$ is not considered in the language sequence, the main problem lies in the inconsistency between $\pi\left(a \mid s_t\right)$ and $\pi\left(a \mid s'_t\right)$. On the one hand, this will lead to a bias in the learning objective of the value model. On the other hand, it is possible that the optimal actor in state $s_t'$ has an extremely low sampling probability in state $s_t$, resulting in the policy being unable to optimize towards the actual optimal direction.

\textbf{Value estimation without the ground truth (Value w/o GT)}

Another choice is to align with the state during policy sampling, the ground truth can be excluded from consideration during value estimation. However, the issue with this method is the information asymmetry between the value model and the reward model. The value model takes $q$ and $y$ as inputs and thus cannot accurately estimate $p(y \succ y^* \mid q)$. A straightforward idea is to distill a pointwise discriminative reward model (abbreviated to pointwise RM) from a pairwise generative reward model (abbreviated to generative RM), and then use this pointwise RM to calculate the optimization objective of the value model. Specifically, given a prompt $q$, the ground truth $y^*$, and the sampled answer $y$, let the score given by the generative RM be denoted as $r(y\mid y*,q)$, which is consistent with the above description. The scores given by the pointwise RM to $y$ and $y^*$ are denoted as $r_{p}(y\mid q)$ and $r_{p}(y^*\mid q)$ respectively. We use MSE loss to distill the pointwise RM:
\begin{equation}
\mathcal{L}_{distill}=\left(r(y\mid y*,q) - \left(r_p(y\mid q)-r_p(y^*\mid q)\right)\right)^2.
\label{eq:xc_1}
\end{equation}
During the process of updating the policy model, we use the reward generated by the generative RM to calculate the advantage $adv$:
\begin{equation}
adv_t=\sum_{l=0}^{\infty}\left(\gamma \lambda\right)^l \delta^{V}_{t+l},\,\, \delta^{V}_t=-V(s_t)+r(y\mid y*,q)+\gamma V(s_{t+1}).
\label{eq:xc_2}
\end{equation}
While during the process of updating the value model, we use the reward generated by the pointwise RM to calculate the return $G$:
\begin{equation}
G_t=\sum_{l=0}^{\infty}\left(\gamma \lambda\right)^l \delta^{V'}_{t+l} + V(s_t),\,\, \delta^{V'}_t=-V(s_t)+r_g(y\mid q)+\gamma V(s_{t+1}).
\label{eq:xc_3}
\end{equation}

\section{Experiments}


\subsection{Setup}
\textbf{Datasets}. To evaluate the effectiveness of our pairwise reward model against traditional point-wise RMs, we conducted experiments on three datasets: an internal Chinese dataset with in-domain IID responses, an external dataset with out-of-distribution (OOD) prompts annotated via internal rules, and the Reward Bench dataset. Moreover, to evaluate the end-to-end effectiveness of pairwise PPO, we conducted experiments on internal and external datasets separately. The internal datasets comprehensively examines seven capabilities of the model, including reasoning and planning, instruction following, STEM, coding, knowledge, fundamental NLP tasks, and long text processing. The external datasets include MMLU pro \cite{mmlu-pro}, CEval \cite{ceval}, KORBench \cite{kor-bench}, BBH \cite{bbh}, MATH \cite{math}, LiveCodeBench \cite{livecodebench}, IFEval \cite{IFEval}, and GPQA \cite{GPQA}.

\textbf{Baseline models}. In our experiments, we compare pairwise reward model with traditional point-wise RM, and compare pairwise PPO with standard PPO \cite{ppo}.
To address computational constraints while maintaining rigorous evaluation, we conducted ablation studies on smaller in-house models to isolate the contributions of individual components and finally run an end to end comparison on larger models.

\textbf{Basic setting}. 
For the comparison between pairwise rm and pointwise rm, both models are trained with a maximum sequence length of \( 16\text{k} \) tokens to accommodate long-form prompts and responses. $\zeta$ is set to be 0.1 and the pairwise RM is optimized with a learning rate scheduler that use cosine decay to decay from \( 5 \times 10^{-6} \) to \( 5 \times 10^{-7} \) over the course of training. We use a batch size of \( 128 \) and train on \( 261,760 \) human labeled pairwise preference comparisons.

For pairwise PPO and baseline PPO training, the learning rates for the policy model and the value model are set to $7.5 \times 10^{-7}$. Benefiting from the robustness of pairwise RM, the policy can be trained to a relatively large KL divergence without the occurrence of reward hacking. Therefore, we set the KL penalty coefficient to $0.001$. The lengths of both the prompt and the response are set to 2k. The $\lambda$ used for GAE is set to $0.95$. Regarding the sampling parameters in the RL training process, we randomly set the top-p value of half of the samples to $0.7$ and that of the other half to $1.0$. The setting of $0.7$ is to align with the sampling parameters during the actual deployment of the model, while the setting of $1.0$ is to expand the sampling space and enhance the exploration of the policy model. 

\subsection{Overall Experimental Results}
To evaluate the performance of the pairwise reward model in the Pairwise-RL framework, we focus on front and back accuracy metrics, comparing them against the accuracy of traditional pointwise RMs.

\begin{table}[htbp]
	\centering
        \caption{Comparison between RMs on smaller size base model}
	\begin{tabular}{|l|ccc|c|}
		\hline
		\multirow{2}{*}{Dataset} & \multicolumn{3}{c|}{Pairwise}                                                                 & Pointwise \\ \cline{2-5}
		                         & Front acc  & \multicolumn{1}{|c}{Back acc} & \multicolumn{1}{|c|}{Avg acc} & Acc       \\ \hline
		IID                      & 0.683          & 0.684                             & \textbf{0.695}                             & 0.658         \\
		OOD               & 0.744          & 0.705                             & \textbf{0.744}                             & 0.709         \\
		Reward Bench             & 0.895          & 0.902                             & 0.910                             & \textbf{0.913}         \\ \hline
	\end{tabular}
	\label{tab:rm_compare}
\end{table}

The experimental results in Table\ref{tab:rm_compare} demonstrate that the pairwise reward model outperforms the traditional pointwise RM on both in-domain IID and out-of-distribution  datasets. On the Reward Bench dataset, which is a relatively simple eval set, the pairwise RM achieves comparable performance to the pointwise RM. Notably, the pairwise RM's average accuracy across front and back evaluations consistently surpasses the individual front and back accuracies, suggesting that combining these evaluations leads to a slight but meaningful improvement in overall performance, likely due to reduced positional bias and enhanced robustness in relative judgments.

To verify the effectiveness of pairwise RL, we conduct RL training on an in-house 70B model. For pairwise RL, we use the ground truth by default when estimating values (Value w/ GT), as its performance is significantly better than the method that does not refer to the ground truth (Value w/o GT). The relevant comparative experiments will be mentioned in the following sections. The experimental results of comparing PPO and Pairwise PPO are shown in Table \ref{tab:comparison1} and Table \ref{tab:comparison2}. 
\begin{table}[htbp]
    \centering
    \caption{Comparison between PPO and Pairwise PPO (Internal dataset)}
    \label{tab:comparison1}
    \begin{tabular}{lcccccccc}
        \toprule
        Model & Reasoning & Instruct & STEM & Coding & Knowledge & NLP & Long-Text & overall \\
        \midrule
        PPO & 48.7 & 47.4 & \textbf{59.2} & 63.4 & 43.3 & 45.2 & 70.6 & 50.9 \\
        Pairwise PPO & \textbf{51.1} & \textbf{49.1} & 58.8 & \textbf{64.2} & \textbf{44.2} & \textbf{48.2} & \textbf{73.1} & \textbf{52.3} \\
        \bottomrule
    \end{tabular}
\end{table}

\begin{table}[htbp]
    \centering
    \caption{Comparison between PPO and Pairwise PPO (External dataset)}
    \label{tab:comparison2}
    \begin{tabular}{lcccccccc}
        \toprule
        Model & MMLU-pro & CEval & KOR & BBH & MATH & LiveCode & IFEval & GPQA \\
        \midrule
        PPO & 74.1 & 88.3 & 50.2 & \textbf{90.5} & 82.5 & \textbf{47.4} & 84.3 & 52.1 \\
        Pairwise PPO & \textbf{74.9} & \textbf{88.4} & \textbf{54.2} & 90.0 & \textbf{82.9} & 45.9 & \textbf{86.7} & \textbf{55.1} \\
        \bottomrule
    \end{tabular}
\end{table}

Compared with the baseline PPO, Pairwise PPO shows significant improvements mainly in reasoning, instruction following, NLP tasks, and long-text tasks on the internal evaluation dataset. On external evaluation datasets, it demonstrates greater enhancements in KOR, IFEval, and GPQA. The improvements in instruction following and NLP-related tasks mainly stem from the more accurate judgments of pairwise RM. As for the enhancements in reasoning-related tasks, they primarily result from a significant alleviation of the reward hacking issue during the RL training process, which enables more comprehensive training. The alleviation of the reward hacking problem is mainly attributed to the utilization of pairwise RL, which scores by comparing the online-generated results with the ground truth. This approach reduces the excessive optimization of the reward model for details that are irrelevant to the actual performance.

\subsection{Model performance under different settings}
In this section, we will compare the performance of pairwise RL under different settings. Considering the experimental costs, all ablation experiments are carried out on a model with a smaller size. The model is evaluated on another internal evaluation dataset. The evaluation dimensions of this dataset include reasoning, math, coding, knowledge, language, and instruction following.
\subsubsection{The influence of the use of ground truth on value estimation}
As discussed in Section \ref{qrl}, there are two ways, Value w/ GT and Value w/o GT, to utilize the ground truth when pairwise PPO conducts value estimation. In order to verify which method yields better results, we conducted a comparative experiment and the experimental conclusions are shown in Table \ref{tab:comparison3}.
\begin{table}[htbp]
    \centering
    \caption{Comparison between Value w/ GT and Value w/o GT on internal datasets}
    \label{tab:comparison3}
    \begin{tabular}{lcccccccc}
        \toprule
        Model & Reasoning & Math & Coding & Knowledge & Language & Instruction \\
        \midrule
        Value w/ GT & \textbf{39.5} & \textbf{36.4} & \textbf{18.9} & \textbf{58.6} & \textbf{59.5} & 38.0 \\
        Value w/o GT & 36.8 & 34.0 & 15.8 & 55.7 & 57.2 & \textbf{38.4} \\
        \bottomrule
    \end{tabular}
\end{table}
According to the experimental results, the performance of Value w/o GT is significantly worse, although this method not only unifies the input sequences of the value model and the policy model, but also aligns the actual meaning of the pairwise RM scoring by subtracting the scores of the roll-out sequence and the ground truth estimated respectively. It is due to the losses existing in the distillation process represented by \ref{eq:xc_2}, the accuracy of the actual value model estimation is not high. There is actually a difficulty in this seemingly simple distillation task: when the pairwise RM scores, it can simultaneously observe the contents of the two responses to form a direct comparison, but the point RM cannot achieve this during the process of predicting the scores of the two responses separately.

\subsection{The influence of the quality and stability of the ground truth}
In order to train the model using pairwise RL, we need to sample a ground truth, which is used as a reference for the pairwise RM, for each prompt in the training set. Ideally, it would be best to obtain the most perfect response through manual annotation. However, this approach is not conducive to the scaling of the RL training set. To ensure that the RL training data does not require annotation intervention, we obtain the ground truth by sampling the best-of-n of the model after supervised fine-tuning, which is also serve as the initialization model for RL training. This method has another advantage: the distribution of the ground truth is consistent with that of the response generated online during the RL training process, which is more conducive to improving the robustness of the pairwise RM scoring.

Intuitively, both the quality and stability of the ground truth will affect the RL training effect. When the quality of the ground truth is high, it can provide sufficient optimization space for RL, enabling the results generated online during the training process to be optimized from "worse than the ground truth" to "better than the ground truth", rather than from "better than the ground truth" to "even better than the ground truth". On the other hand, stability means that the ground truth should maintain a stable level as much as possible under different prompts, so that the scoring meaning of the pairwise RM remains stable. In order to verify the effects of these two variables on RL training, we conducted multiple groups of experiments with different values of $n$ when sampling the best-of-n, and the results are shown in Table \ref{tab:comparison4}.

\begin{table}[htbp]
    \centering
    \caption{Model performance under different quality and stability of ground truth}
    \label{tab:comparison4}
    \begin{tabular}{lccccccc}
        \toprule
        Ground truth & Reasoning & Math & Coding & Knowledge & Language & Instruction & Overall\\
        \midrule
        worst of 5 & 40.6 & 36.2 & 16.6 & 57.3 & 59.5 & 37.5 & 40.1 \\
        best of 1 & 37.4 & 34.5 & 18.6 & 58.2 & 57.7 & 37.8 & 39.2 \\
        best of 5 & 41.0 & 37.0 & 16.6 & 58.4 & 61.0 & 38.4 & 40.7 \\
        best of 10 & 40.3 & 36.7 & 15.9 & 59.8 & 60.8 & 39.8 & 41.1 \\
        best of 15 & 42.9 & 37.2 & 16.7 & 59.2 & 62.2 & 37.3 & 41.4 \\
        best of 30 & 43.3 & 37.3 & 17.3 & 60.5 & 62.5 & 38.7 & 41.7 \\
        best of 40 & 40.4 & 37.7 & 16.1 & 59.2 & 61.1 & 38.8 & 41.3 \\
        best of 50 & 39.9 & 38.2 & 16.7 & 58.9 & 61.6 & 39.8 & 41.2 \\
        \bottomrule
    \end{tabular}
\end{table}

The experimental results show that within a certain range, as the quality of the ground truth gradually improves (from worst of 5 -> best of 5 -> best of 10 -> best of 15 -> best of 30), the performance of the final model training will gradually increase. However, when $n$ continues to increase (best of 40, best of 50), the performance of the model starts to decline. The reason is that a pointwise RM is used when selecting the best-of-n. Therefore, as n increases, the accuracy of the RM will also decrease. When $n$ exceeds 30, the higher scores given by the RM may be untrustworthy. In addition, when the ground truth is selected as the worst of 5, the effect is actually better than when it is selected as the best of 1. The reason is that the ground truth obtained by sampling the best of 1 has poor stability. The quality of the ground truth varies under different prompts, which will lead to the instability of the scoring meaning of the pairwise RM. 

\begin{figure}[h]
    \centering
    \begin{subfigure}[b]{0.45\textwidth}
        \centering
        \includegraphics[width=\textwidth]{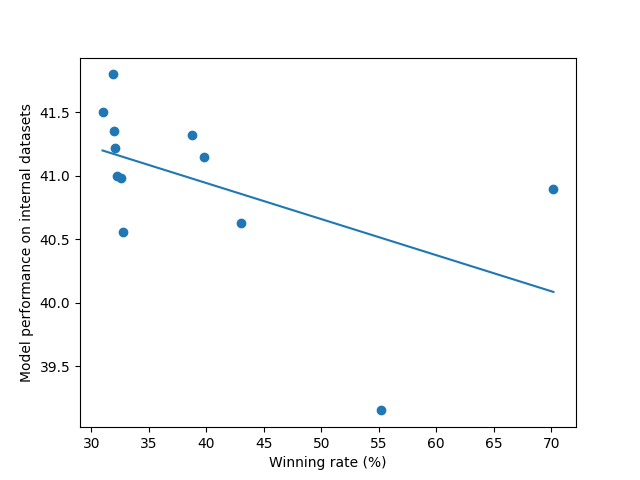}
        \label{fig:first}
    \end{subfigure}
    \hfill
    \begin{subfigure}[b]{0.45\textwidth}
        \centering
        \includegraphics[width=\textwidth]{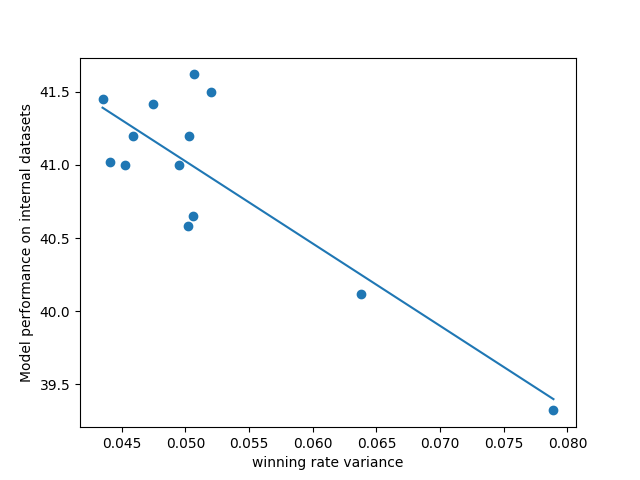}
        \label{fig:second}
    \end{subfigure}
    \caption{The effect of the quality and stability of ground truth on the model's performance}
    \label{fig:both}
\end{figure}

Furthermore, we employ two quantitative indicators to measure the quality and stability of the ground truth. To assess the quality, we use pairwise RM to calculate the \textbf{winning rate} of the SFT model best-of-1 response against the ground truth. A lower winning rate indicates better quality of the ground truth. To measure the stability, we calculate the winning rates of the ground truth for all prompts in a subset of training data and then compute the \textbf{winning rate variance}. A larger variance implies poorer stability of the ground truth. We conduct multiple experiments with different selection strategies for ground truth and the results are shown in Figure \ref{fig:both}. According to the experimental results, it is easy to observe that as the quality or stability of the ground truth deteriorates, the final performance of the model will also decline. 

\section{Conclusion}
In this paper, we introduce Pairwise-RL, a novel RLHF framework designed to address two critical limitations in traditional RLHF approaches: the calibration challenges of scalar rewards derived from pairwise comparisons and the mismatch between generative model initialization and discriminative reward modeling tasks. Pairwise-RL unifies reward model training and reinforcement learning within a consistent pairwise paradigm, leveraging generative modeling techniques to enhance reward model performance and calibration. Specifically, we propose a pairwise reward model that directly evaluates response pairs, mitigating positional biases through data augmentation and symmetric loss constraints, and a pairwise PPO algorithm that maximizes the win probability of generated responses over ground-truth anchors. Experimental results demonstrate that Pairwise-RL outperforms conventional RLHF methods on both internal and external benchmarks, achieving improved alignment with human preferences and enhancing model behavior across diverse tasks, including reasoning, instruction following, and long-text processing. This work underscores the effectiveness of a unified pairwise approach in overcoming the limitations of traditional RLHF, offering a robust framework for aligning large language models with human values.

\clearpage

\bibliographystyle{plainnat}
\bibliography{main}

\clearpage

\section{Appendix}
We show here that decay excessively large advantages serves to reduce the upper bound of the KL divergence between the policy before and after the update on these samples. When training RL model, the gradient update rule is given by: 
\begin{equation}
\theta_{\text{new}} = \theta_{\text{old}} + \alpha \nabla_{\theta}J(\theta_{\text{old}}),
\label{eq:gradient_update}
\end{equation}
where \(\alpha\) is the step size. To analyze the KL divergence \(\text{KL}(\pi_{\theta} || \pi_{\theta_{\text{old}}})\), we use a second-order expansion around \(\theta_{\text{old}}\), which approximates the KL divergence as: 
\begin{equation}
\text{KL}(\pi_{\theta} || \pi_{\theta_{\text{old}}}) \approx \frac{1}{2} (\theta - \theta_{\text{old}})^T F(\theta_{\text{old}}) (\theta - \theta_{\text{old}}),
\label{eq:kl_expand}
\end{equation}
where \(F(\theta)\) is the Fisher information matrix defined as:
\begin{equation}
F(\theta) = \mathbb{E}_{a \sim \pi_{\theta}} \left[ \nabla_{\theta} \log \pi_{\theta}(a|s) \nabla_{\theta} \log \pi_{\theta}(a|s)^T \right].
\label{eq:Fisher}
\end{equation}
This matrix \(F(\theta)\) corresponds to the covariance of the policy gradient \(\nabla_{\theta} \log \pi_{\theta}(a|s)\). To see this, note that the covariance is:
\begin{equation}
\text{Cov}(\nabla_{\theta} \log \pi_{\theta}(a|s)) = \mathbb{E} \left[ (\nabla_{\theta} \log \pi_{\theta}(a|s))(\nabla_{\theta} \log \pi_{\theta}(a|s))^T \right] - \mathbb{E} \left[ \nabla_{\theta} \log \pi_{\theta}(a|s) \right] \mathbb{E} \left[ \nabla_{\theta} \log \pi_{\theta}(a|s) \right]^T,
\label{eq:covariance}
\end{equation}
the second term vanishes because in our case \(\sum_a \pi_{\theta}(a|s) = 1\) implies \(\mathbb{E}_{a \sim \pi_{\theta}} \left[ \nabla_{\theta} \log \pi_{\theta}(a|s) \right] = 0\). Thus, \(\text{Cov}(\nabla_{\theta} \log \pi_{\theta}(a|s)) = F(\theta)\). Now we have $F$ to be a symmetric positive semi-definite matrix.

Next, consider the policy gradient \(\nabla_{\theta}J(\theta_{\text{old}})\), which is: 
\begin{equation}
\nabla_{\theta}J(\theta_{\text{old}}) = \mathbb{E}_{a \sim \pi_{\theta_{\text{old}}}} \left[ \nabla_{\theta} \log \pi_{\theta_{\text{old}}}(a|s) A(s,a) \right],
\label{eq:pg_eq}
\end{equation}
where \(A(s,a)\) is the advantage function. Applying the Cauchy-Schwarz inequality to the norm of this gradient, we get: 
\begin{equation}
\left\| \nabla_{\theta}J(\theta_{\text{old}}) \right\|^2 \leq \mathbb{E}_{a \sim \pi_{\theta_{\text{old}}}} \left[ A(s,a)^2 \right] \mathbb{E}_{a \sim \pi_{\theta_{\text{old}}}} \left[ \left\| \nabla_{\theta} \log \pi_{\theta_{\text{old}}}(a|s) \right\|^2 \right].
\label{eq:Cauchy-Schwarz}
\end{equation}
The second expectation on the right-hand side is the trace of the Fisher matrix \(F(\theta_{\text{old}})\), since: 
\begin{equation}
\mathbb{E}_{a \sim \pi_{\theta}} \left[ \nabla_{\theta} \log \pi_{\theta}(a|s)^T \nabla_{\theta} \log \pi_{\theta}(a|s) \right] = \text{Tr}(F(\theta)).
\label{eq:trace_f}
\end{equation}
Thus, the inequality becomes: 
\begin{equation}
\left\| \nabla_{\theta}J(\theta_{\text{old}}) \right\|^2 \leq \mathbb{E}_{a \sim \pi_{\theta_{\text{old}}}} \left[ A(s,a)^2 \right] \text{Tr}(F(\theta_{\text{old}})).
\label{eq:norm_ineq}
\end{equation}
Since \(F(\theta_{\text{old}})\) is symmetric and positive semi-definite, we use the property that for any vector \(v\), \(v^T F v \leq \text{Tr}(F) \|v\|^2\). Applying this to \(v = \nabla_{\theta}J(\theta_{\text{old}})\), we obtain: 
\begin{equation}
\nabla_{\theta}J(\theta_{\text{old}})^T F(\theta_{\text{old}}) \nabla_{\theta}J(\theta_{\text{old}}) \leq \text{Tr}(F(\theta_{\text{old}})) \left\| \nabla_{\theta}J(\theta_{\text{old}}) \right\|^2.
\label{eq:kl_ineq1}
\end{equation}
Substituting the earlier bound on \(\left\| \nabla_{\theta}J(\theta_{\text{old}}) \right\|^2\) into this inequality gives: 
\begin{equation}
\nabla_{\theta}J(\theta_{\text{old}})^T F(\theta_{\text{old}}) \nabla_{\theta}J(\theta_{\text{old}}) \leq \mathbb{E}_{a \sim \pi_{\theta_{\text{old}}}} \left[ A(s,a)^2 \right] \left( \text{Tr}(F(\theta_{\text{old}})) \right)^2.
\label{eq:kl_ineq2}
\end{equation}

Finally, substituting this result back into the KL divergence approximation, we find: 
\begin{equation}
\text{KL}(\pi_{\theta} || \pi_{\theta_{\text{old}}}) \approx \frac{\alpha^2}{2} \nabla_{\theta}J(\theta_{\text{old}})^T F(\theta_{\text{old}}) \nabla_{\theta}J(\theta_{\text{old}}) \leq \frac{\alpha^2}{2} \mathbb{E}_{a \sim \pi_{\theta_{\text{old}}}} \left[ A(s,a)^2 \right] \left( \text{Tr}(F(\theta_{\text{old}})) \right)^2.
\label{eq:kl_bound}
\end{equation}

This establishes the upper bound on the KL divergence in terms of the step size \(\alpha\), the expected squared advantage, and the trace of the Fisher information matrix. Decaying excessively large advantages reduces the term \(\mathbb{E}_{a \sim \pi_{\theta_{\text{old}}} \left[ A(s,a)^2 \right]}\), thereby decreasing the upper bound on the KL divergence.


\end{document}